\documentclass{article}

\usepackage{arxiv}

\usepackage[utf8]{inputenc} % allow utf-8 input
\usepackage[T1]{fontenc}    % use 8-bit T1 fonts
\usepackage{hyperref}       % hyperlinks
\usepackage{url}            % simple URL typesetting
\usepackage{booktabs}       % professional-quality tables
\usepackage{amsfonts}       % blackboard math symbols
\usepackage{nicefrac}       % compact symbols for 1/2, etc.
\usepackage{microtype}      % microtypography
\usepackage{lipsum}
\usepackage{tikz}
\usepackage{graphicx}
\usepackage{multirow}
\usepackage{amsmath}
\usepackage{newclude}
\usepackage{todonotes}
\usepackage{subcaption}
\usepackage[symbol]{footmisc}

\title{COVID-19 Prognosis via Self-Supervised Representation Learning and Multi-Image Prediction}

\author{
    Anuroop Sriram$^*$\\
    Facebook AI Research
    \And
    Matthew Muckley$^*$\\
    Facebook AI Research
    \And
    Koustuv Sinha\\
    Facebook AI Research
    \And
    Farah Shamout\\
    NYU Abu Dhabi
    \And
    Joelle Pineau\\
    Facebook AI Research
    \And
    Krzysztof J. Geras\\
    NYU School of Medicine
    \And
    Lea Azour\\
    NYU School of Medicine
    \And
    Yindalon Aphinyanaphongs\\
    NYU School of Medicine
    \And
    Nafissa Yakubova\\
    Facebook AI Research
    \And
    William Moore\\
    NYU School of Medicine\\
}

\begin{document}
\maketitle

\begin{abstract}
The rapid spread of COVID-19 cases in recent months has strained hospital resources, making rapid and accurate triage of patients presenting to emergency departments a necessity.
Machine learning techniques using clinical data such as chest X-rays have been used to predict which patients are most at risk of deterioration.
We consider the task of predicting two types of patient deterioration based on chest X-rays: adverse event deterioration (i.e., transfer to the intensive care unit, intubation, or mortality) and increased oxygen requirements beyond 6 L per day.
Due to the relative scarcity of COVID-19 patient data, existing solutions leverage supervised pretraining on related non-COVID images, but this is limited by the differences between the pretraining data and the target COVID-19 patient data.
In this paper, we use self-supervised learning based on the momentum contrast (MoCo) method in the pretraining phase to learn more general image representations to use for downstream tasks. We present three results.
The first is deterioration prediction from a single image, where our model achieves an area under receiver operating characteristic curve (AUC) of 0.742 for predicting an adverse event within 96 hours (compared to 0.703 with supervised pretraining) and an AUC of 0.765 for predicting oxygen requirements greater than 6 L a day at 24 hours (compared to 0.749 with supervised pretraining).
We then propose a new transformer-based architecture that can process sequences of multiple images for prediction and show that this model can achieve an improved AUC of 0.786 for predicting an adverse event at 96 hours and an AUC of 0.848 for predicting mortalities at 96 hours.
A small pilot clinical study suggested that the prediction accuracy of our model is comparable to that of experienced radiologists analyzing the same information.\footnotetext[1]{Equal contribution.}
\end{abstract}

\section{Introduction}
\label{sec:intro}
\include*{intro}
The SARS COV-2 infection (COVID-19) has caused a surge of patients with respiratory illness presenting to emergency departments worldwide throughout 2020. 
When patients arrive at the emergency department, hospitals need to evaluate their risk of deterioration for effective clinical decision making and resource allocation. This is especially important during a pandemic when hospital resources are strained.
Initial patient assessment often includes imaging studies, and these initial assessments can be used for resource planning within the hospital.
Chest X-ray radiography (CXR) is a well-established imaging examination for patients with symptoms of shortness of breath and fever \cite{bhargavan2005utilization}.
Although traditionally used for diagnosis, the use of CXR has expanded to COVID deterioration prediction \cite{shamout2020artificial,Kwon2020CombiningIR,zhang2020clinically} and is a standard component of COVID patient health assessments.

Deep learning methods for image-based diagnosis are a standard tool used in radiology research \cite{cheng2016computer,rajpurkar2017chexnet,chartrand2017deep,lakhani2017deep,yamashita2018convolutional,rajpurkar2018deep} and have been applied to COVID-19 \cite{shamout2020artificial,Kwon2020CombiningIR,chen2020momentum,amyar2020multi,zhang2020diagnosis,oh2020deep,caron2020unsupervised}.
Most deep learning methods for radiology rely on the collection of large numbers of labeled radiology images for supervised training, which introduces a number of constraints.
First, the collection of large, labeled, training sets is expensive \cite{chartrand2017deep}, and can only be accomplished by well-funded research organizations.
Second, it can be difficult to assign labels for many radiology tasks. Radiologists are often tasked to give impressions of images and radiology reports are expected to convey nuance and uncertainty with regards to the imaging findings.
This can create downstream uncertainty in label assignment \cite{irvin2019chexpert}. Third and related, the requirement for large datasets often restricts the use of deep learning methods to established and well-understood pathologies. Thus far, large public X-ray datasets have been collected at major research centers over time spans as long as a decade \cite{wang2017chestx,irvin2019chexpert,johnson2019mimic}. The availability of such data is essential to facilitate new research into machine learning methods for improving image-based diagnosis. However these datasets do not capture data from emerging diseases, such as COVID-19.

At this time, most hospital centers have not collected enough COVID-19 patient data at the scale required to train deep learning representations.
Technological and privacy considerations prevent data pooling between many research centers.
One common approach to dealing with small amounts of labeled data is to apply transfer learning \cite{shin2016deep}, where a model is pretrained in a supervised fashion on a large, labeled dataset (e.g., ImageNet), then finetuned on the task of interest.
However, transfer learning can lead to poor performance if the tasks are too different, and the model is not able to learn the features necessary for the transfer task during the pretraining step.

Recently, new self-supervised methods which rely on contrastive losses have been shown to generate representations that are as good for classification as those generated using purely supervised methods \cite{hadsell2006dimensionality,chen2020simple,he2020momentum,chen2020improved}.
The advantage of contrastive loss functions is that they are able to achieve feature extraction independent of labels or tasks associated with the pretraining dataset.
These features are then used for training a classifier in the fine-tuning stage with the target data.
In this paper, we study the applicability of self-supervised learning to the task of COVID-19 deterioration prediction.
We pretrain a model using momentum contrast (MoCo) \cite{he2020momentum,chen2020improved} on two large, public chest X-ray datasets,  MIMIC-CXR-JPG \cite{johnson2019mimic,johnson2019mimicjpg,goldberger2000physiobank} and CheXpert \cite{irvin2019chexpert}, and then use the pretrained model as a feature extractor for the downstream task of prediction COVID-19 patient outcomes. Concurrent to our work, Sowrirajan et al. \cite{sowrirajan2020moco} have shown that MoCo pre-training can be used to learn representations from chest X-rays that are useful for identifying the presence of pleural effusions. 

We study the effectiveness of this approach on three downstream tasks: 1) adverse event prediction from single images (SIP), oxygen requirements prediction from single images (ORP), and adverse event prediction from multiple images (MIP).
We find that, when using the DenseNet architecture \cite{huang2017densely,rajpurkar2017chexnet,rajpurkar2018deep}, self-supervised training achieved higher areas under receiver operating characteristic curve (AUC) on a hold-out test set than supervised training for predicting an adverse event at all time points.
Self-supervised pretrained models achieved higher AUC values for the oxygen requirements prediction task at early time points.
We then show that performance for predicting adverse events can be further improved by making predictions based on image sequences with a Transformer model.
Finally, we characterize the performance of our single-image models vs. the previous COVID-GMIC model \cite{shamout2020artificial} and our multi-image models vs. human radiologists in a pilot clinical study.
In combination with this paper, we also open source our code and models pretrained on public data at \url{https://github.com/facebookresearch/CovidPrognosis} for other researchers interested in using them for fine-tuning on their own X-ray datasets.

\section{Datasets}
\label{sec:dataandtasks}

\subsection{Public Chest X-ray Datasets (Pretraining Data)}

We used two datasets for pretraining our models~\cite{johnson2019mimic,johnson2019mimicjpg,goldberger2000physiobank} and CheXpert \cite{irvin2019chexpert}.
The MIMIC-CXR dataset consists of 377,110 chest X-ray images corresponding to 227,835 radiographic studies performed at the Beth Israel Deaconess Medical Center in Boston, MA.
The CheXpert dataset consists of 224,316 chest radiographs of 65,240 patients compiled from chest radiographic studies performed at the Stanford Hospital.
The datasets did not contain any COVID-19 patients during the time of our study as they were collected before the COVID-19 pandemic.
Imaging findings included a variety of chest conditions such as atelectasis, cardiomegaly, consolidation, edema, and pleural effusion, based on a labeler designed for the CheXpert dataset \cite{johnson2019mimic,irvin2019chexpert}.
We did not use these findings in our study beyond initial validation of the training pipeline.

\subsection{COVID-19 Dataset and Deterioration Labels (Fine-tuning Data)}

For fine-tuning, we used an extended version of the NYU COVID dataset previously described \cite{shamout2020artificial}, containing 26,838 X-ray radiographs from 4,914 patients.
COVID-19 prognosis can be defined by a wide array of factors.
Our goal in this study is to predict patient deterioration that will require physical transfers throughout the hospital or altered equipment usage.
As an example, one type of deterioration would be a transfer to the ICU.
During the transfer process, a COVID-positive patient is removed from isolation, heightening the risk of intra-hospital infection.
By locating at-risk patients closer to the ICU from the beginning, the the transfer time can be reduced, correspondingly reducing the risk of intra-hospital infection. 
Additionally, there and limited number of intensive care beds available per hospital.  
Predicting the need for these beds is critical to hospital resource management. 
Another form of deterioration is increased oxygen utilization.
Normally, patients are administered oxygen via the nasal cannula.
When a patient's oxygen requirements exceeds 6 L per day, oxygen administration through the nasal cannula is no longer possible and delivery must be done via a mask.

Formally, our study considers two label classes: 1) adverse events \cite{shamout2020artificial,Kwon2020CombiningIR} and 2) increased oxygen requirements.
An ``adverse event'' consists of any of the three events: transfer to the intensive care unit (ICU), intubation, or mortality.
We labeled each image with whether the patient developed any adverse event within 24, 48, 72 or 96 hours of the scan.
We obtained these labels by occurrences of adverse events recorded in the patient's anonymized electronic health records data shared by NYU, following previous practice \cite{shamout2020artificial}.
We define ``increased oxygen utilization'' as an event where a patient requires more than 6 L of oxygen in a day.
Similarly to the case of adverse events, we labeled each image in the dataset with whether the patient required increased oxygen on within 24, 48, 72, or 96 hours of the scan.

\subsection{Definition of Prediction Tasks}
\label{subsec:predtasks}

The goal of our machine learning models is to predict deterioration labels from either single chest radiograph or sequences of chest radiographs.
The first task is to predict adverse events from a single chest radiograph.
This constitutes a multi-class classification setting for ICU transfer, intubation, mortality, and ``Any'' adverse event at 24, 48, 72, 96, and ``Any'' time windows (20 total labels).
We call this task the \emph{Single Image Prediction (SIP)} task.
The second task is to predict increased oxygen requirements from a single chest radiograph.
This also constitutes a multi-class classification setting with increased oxygen at 24, 48, 72, 96, and ``Any'' time windows (5 total labels).
We call this the \emph{Oxygen Requirement Prediction (ORP)} task.

Our final task is to predict adverse events from a sequence of radiographs.
This task is of particular interest, as it more closely aligns with how radiologists interpret images.
Radiologists use relative changes in images to identify a patient's trajectory.
For this task, we considered the same multi-class classification problem as in the SIP setting.
We call this task the \emph{Multiple Image Prediction (MIP)} task.

\subsection{Data Splits}

Table \ref{tab:dataset_summary} shows the counts of scans and patients from the NYU COVID data in each split for each task.
The tasks vary in counts due to varying selection criteria used for each task.
For the SIP task, we followed the previous standard \cite{shamout2020artificial} in including chest radiographs of patients collected in the emergency department and excluding chest radiographs of patients who had already experienced any of the three adverse events, as they had already deteriorated.
This enables us to directly compare our results to those in \cite{shamout2020artificial}.
For the ORP task, we only included chest radiographs of patients collected in the emergency department, and did not apply any other exclusion criteria because we considered the oxygen requirement prediction to be independent of the occurrence of any adverse event. 
In the MIP task, the goal was to predict deterioration of patients in the emergency department and during any subsequent hospital admissions.
Therefore, for this task we only excluded chest radiographs collected after a patient had experienced any adverse event.
\begin{table}[htb]
  \centering
  \caption{Subsplits for prediction tasks.}
  \begin{tabular}{lrrrrrrr}
    \toprule
    Task & \multicolumn{3}{c}{\# Scans} & & \multicolumn{3}{c}{\# Patients} \\
        \cmidrule{2-4}\cmidrule{6-8}
         & \{train, val\} & test & total & & \{train, val\} & test & total \\
    \midrule
    SIP & 5,617 & 770 & 6,387 & & 2,943 & 718 & 3,661 \\
    MIP & 17,915 & 2,205 & 20,120 & & 2,774 & 438 & 3,212 \\
    ORP & 15,370 & 5,711 & 21,081 & & 2,887 & 1,096 & 3,983 \\

    % SIP (\# Scans) & 5,617 & 770 & 6387 \\
    % SIP (\# Patients) & 2,943 & 718 & 3,661 \\
    % MIP (\# Scans) & 17,915 & 1,102 & 19017 \\
    % MIP (\# Patients) &  &  & 4,799 \\
    % ORP (\# Scans) & 15,370 & 805 & 16,175 \\
    % ORP (\# Patients) & 2,887 & 256 & 3,143 \\
    \midrule
  \end{tabular}
  \label{tab:dataset_summary}
\end{table}
\begin{figure}[htb]
    \centering
    \includegraphics[scale=0.8]{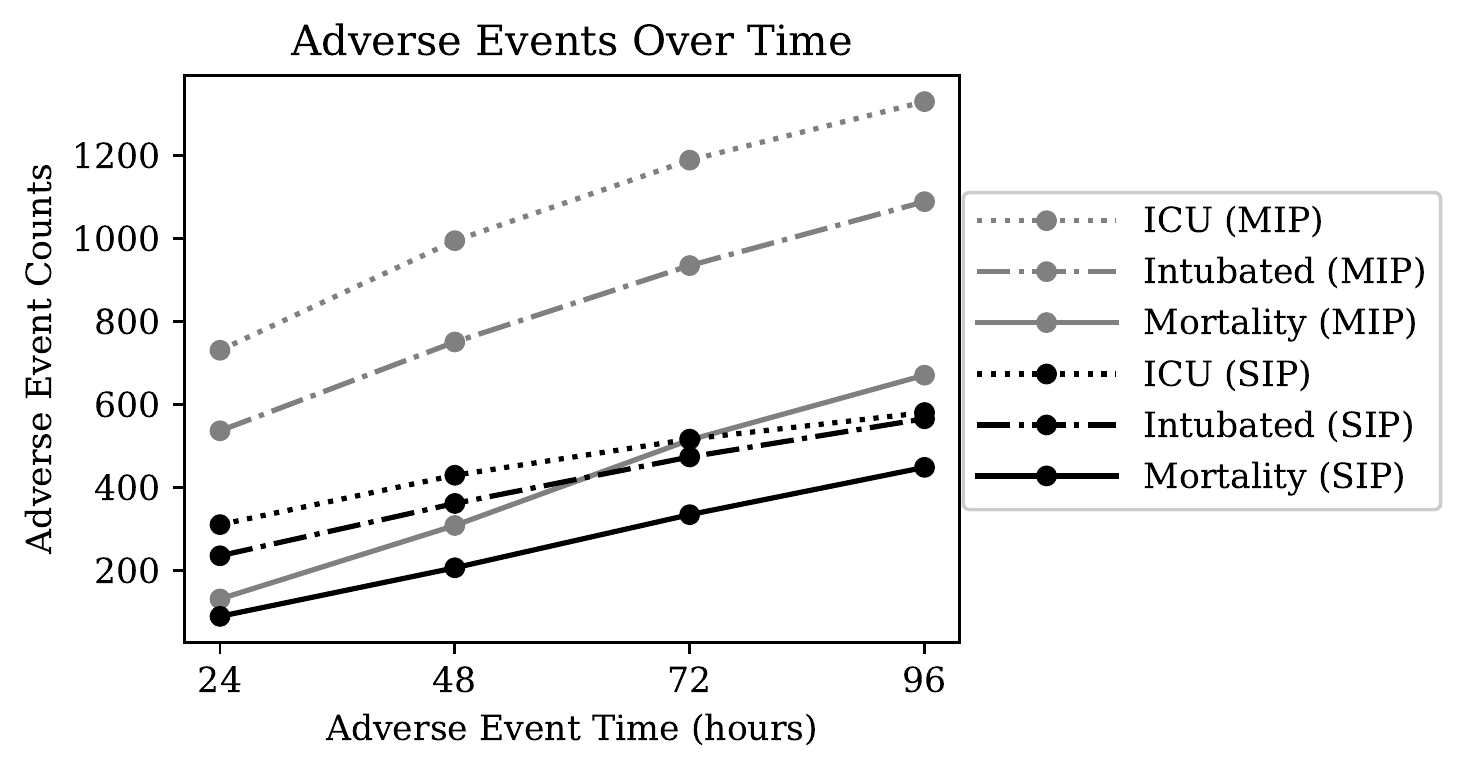}
    \caption{Distribution of adverse events with respect to the time from chest X-ray acquisition for the single-image (SIP) and multiple image (MIP) prediction tasks.}
    \label{fig:adverse_event_labels}
\end{figure}

Figure \ref{fig:adverse_event_labels} shows the number of adverse events in the SIP and MIP splits as a function of time.
We note the labels are necessarily monotonically increasing due to the fact that adverse events at 48 hours include all events at 24 hours, etc.
We did not make any special adjustments to our models to take advantage of this aspect, but this could be a topic for future investigation.

\section{Methods}
\label{sec:models}
Here we describe the models and training process.
All of our proposed models were based on the DenseNet-121 architecture \cite{huang2017densely}, which has been previously applied to chest radiographs \cite{rajpurkar2017chexnet,rajpurkar2018deep,Kwon2020CombiningIR}.
We begin by describing our self-supervised pretraining procedure, then follow with the fine-tuning procedure for the SIP and ORP tasks.
Lastly, we describe the development of a new transformer based architecture that we applied to the MIP task.

\subsection{Self-Supervised Pretraining using Momentum Contrast Learning}
\label{subsec:selfsup_pretraining}

\begin{figure}[htb]
    \centering
    \begin{tikzpicture}
        \node[inner sep=0pt] (base_image) at (0,0)
            {\includegraphics[width=.1\textwidth]
            {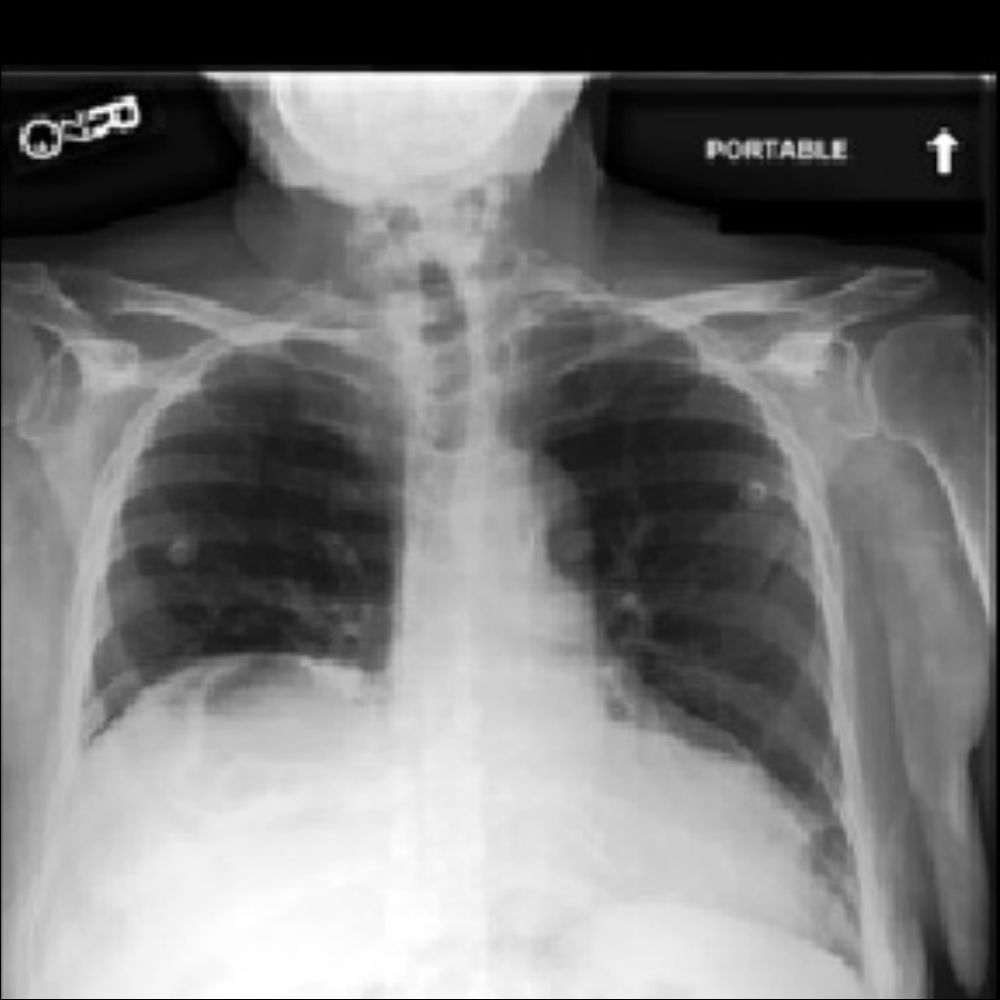}};
        \node[fill=white] (aug_1_title) at (0,-1.2)
            {base image};
            
        % upper path
        \node[inner sep=0pt] (augmentation_1) at (3,1.5)
            {\includegraphics[width=.1\textwidth]
            {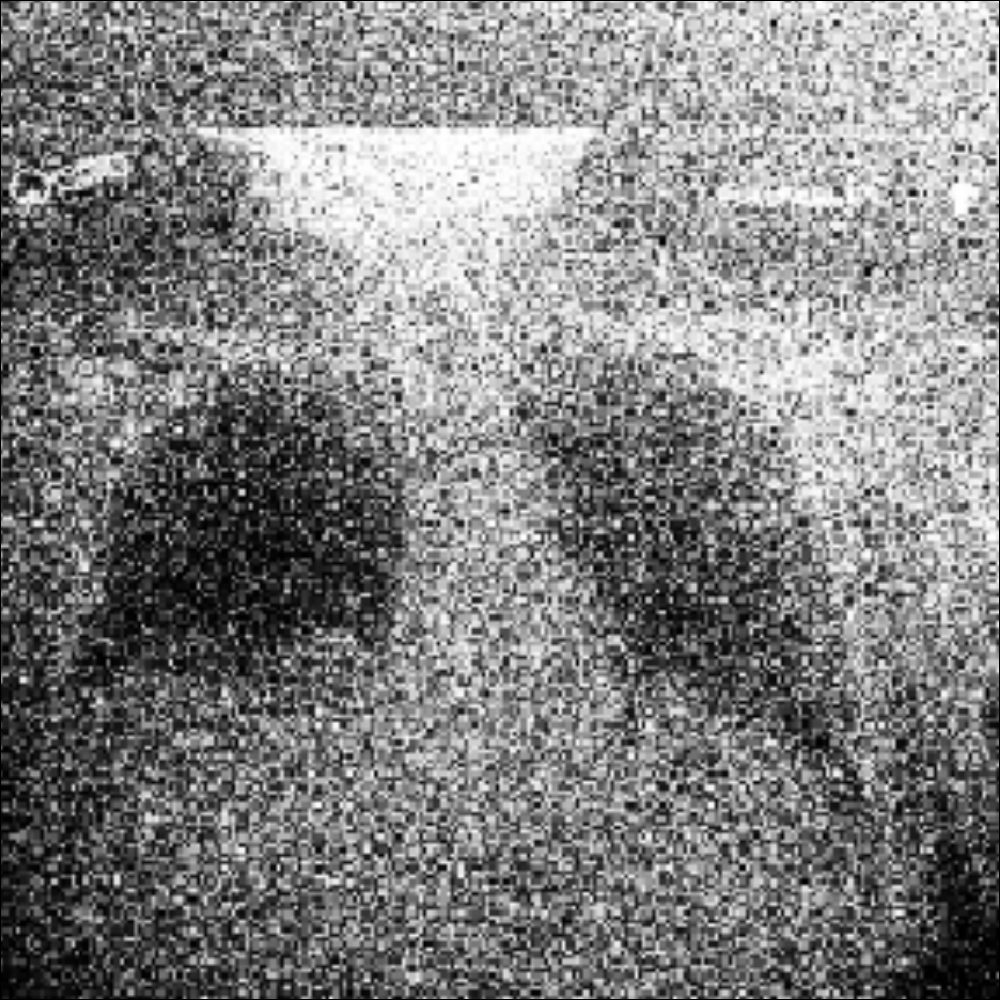}};
        \node[fill=white] (aug_1_title) at (3,0.3)
            {$x_q$};
        \node[fill=black!10!green,rounded corners=2,minimum width=50,minimum height=25]
            (encoder) at (6,1.5) {encoder};
        \node[]
            (rep_q) at (8,1.5) {$r_q$};
            
        % lower path
        \node[inner sep=0pt] (augmentation_2) at (3,-1.5)
            {\includegraphics[width=.1\textwidth]
            {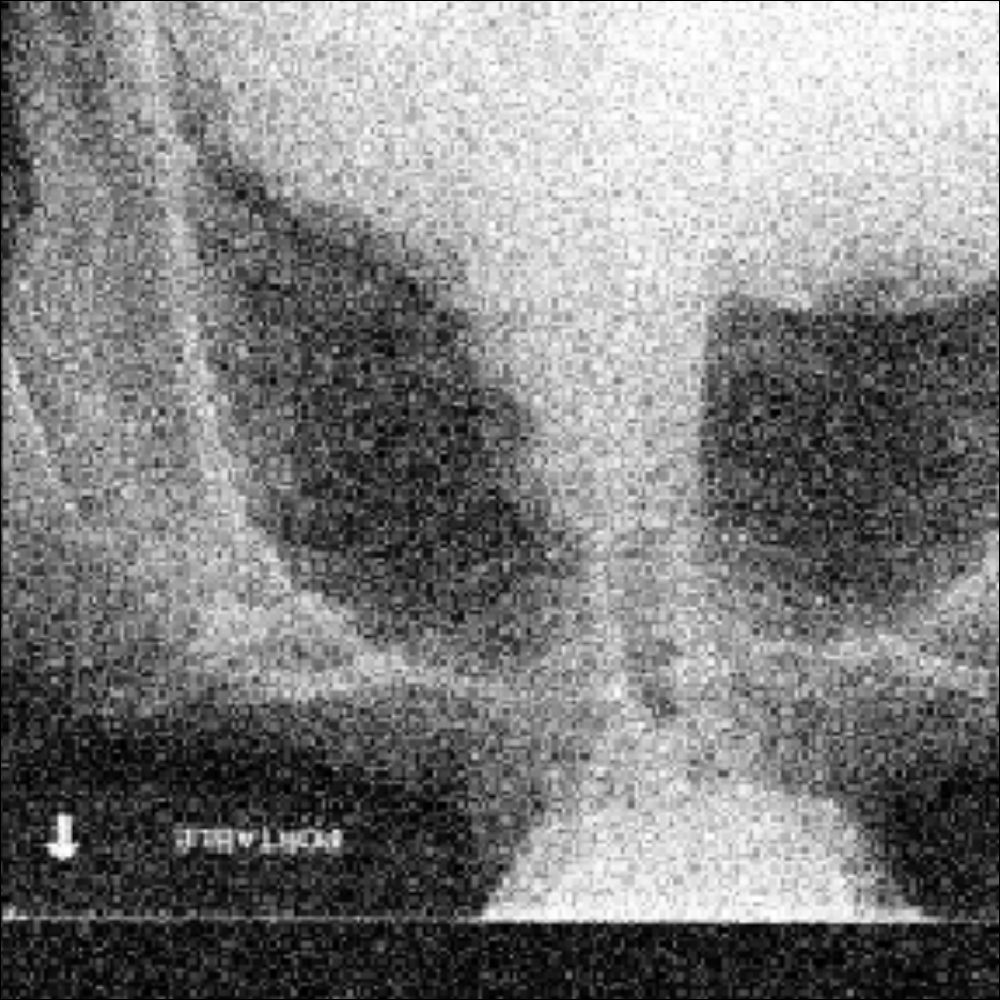}};
        \node[fill=white] (aug_2_title) at (3, -2.7)
            {$x_k$};
        \node[fill=white!80!blue,rounded corners=2,minimum width=50,minimum height=25,
            text width=50,align=center]
            (mo_encoder) at (6,-1.5) {\centering momentum encoder};
        \node[]
            (rep_k) at (8,-1.5) {$r_k$};
            
        % combine
        \node[]
            (qdotr) at (9,0) {$\mathcal{L}(r_q, r_k)$};
            
        % upper lines
        \draw[->,thick] (base_image.east) -- (augmentation_1.west)
            node[midway,fill=white] {Aug. 1};
        \draw[->,thick] (augmentation_1.east) -- (encoder.west);
        \draw[->,thick] (encoder.east) -- (rep_q.west);
        \draw[->,thick] (rep_q.east) -- +(2em, 0) -- (qdotr.north);
        
        % lower lines
        \draw[->,thick] (base_image.east) -- (augmentation_2.west)
            node[midway,fill=white] {Aug. 2};
        \draw[->,thick] (augmentation_2.east) -- (mo_encoder.west);
        \draw[->,thick] (mo_encoder.east) -- (rep_k.west);
        \draw[->,thick] (rep_k.east) -- +(2em, 0) -- (qdotr.south);
    \end{tikzpicture}
    \caption{Diagram for momentum contrast training. A base image is transformed via two random augmentations (Aug.~1 and Aug.~2) into images $x_q$ and $x_k$. $x_q$ is passed through an encoder network, while $x_k$ is passed through a momentum encoder network. The representations generated by each network are then passed into a contrastive loss that promotes similarity between the representations $r_q$ and $r_k$.}
    \label{fig:contrastive_loss_diagram}
\end{figure}
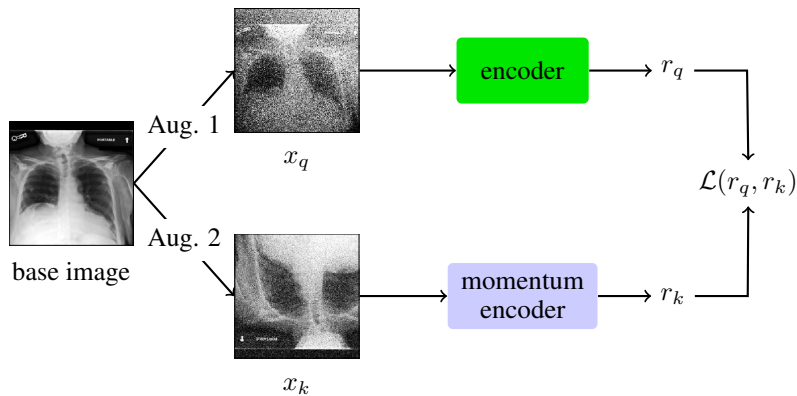

The contrastive loss framework is as follows: a deep neural network is constructed for mapping an image to a latent space \cite{hadsell2006dimensionality}.
The neural network is trained to minimize a contrastive loss.
The contrastive loss is constructed such that similar images are mapped to vectors that are closer to each other (as measured by a contrastive loss function) while dissimilar images are mapped to vectors that are further apart.
Figure \ref{fig:contrastive_loss_diagram} shows an schematic of the contrastive loss training procedure.

Each training step begins by selecting a base image, $x$, from a training dataset of unlabeled images.
Then, two different augmentations are chosen at random and applied to the base image separately to generate two augmented images, the query image $x_q$ and the key image $x_k$. These images are passed through two different neural networks, called the encoder and the momentum encoder respectively, to generate representation $r_q$ for $x_q$ and $r_k$ for $x_k$.

The goal of the contrastive loss is to identify that $r_q$ and $r_k$ come from the same underlying image in the presence of divergent augmentations.
In practice contrastive losses usually require very large batch sizes due to the fact that many negative examples (i.e., values for $r_{q,j}$ and $r_{k,i}$ where $r_{k,i}$ is not from the same image) are necessary to achieve strong performance \cite{chen2020simple}.

MoCo \cite{he2020momentum} is a recent contrastive loss method that avoids the need for large batch sizes by maintaining a queue of representations.
Within the queue, the model stores $K$ examples of $r_{k,i}$ for $i \in [1, ..., K]$.
The model is then asked to identify which of the $K$ examples is the matching one for $r_q$.
This can be modeled mathematically via the InfoNCE contrastive loss function \cite{oord2018representation}:
\begin{equation}
\label{eq:contrast_loss}
    \mathcal{L} \left (r_q, r_k \right ) = - \text{log} \frac{\text{exp}\left (r_q \cdot r_{k+} / \tau \right )}{\sum_{i=1}^K \text{exp}\left (r_q \cdot r_{k,i} / \tau \right )},
\end{equation}
where $\tau$ is a temperature hyperparameter and $K$ is the number of currently stored representations.
The gradient from the contrastive loss is backpropagated to the encoder network, and then the momentum encoder is updated via a momentum update \cite{he2020momentum}.
The momentum update forces the momentum encoder to change more slowly than the encoder network, which helps stabilize training.

\subsection{Single Image Prediction (SIP) and Oxygen Requirement Prediction (ORP) Models}
\label{subsec:sip_model}

The SIP and ORP tasks were described in Section \ref{subsec:predtasks}.
We extend the MoCo encoder model to an image classifier model by appending a linear classifier to the end of the encoder pipeline.
We then fine-tune the model on the COVID image dataset, reshaping input images to a $224 \times 224$ grid and applying random horizontal and vertical flipping during training.
We used the Adam optimizer \cite{kingma2015adam} with a cosine annealing learning rate decay \cite{Loshchilov2017SGDRSG}.

\subsection{Multiple Image Prediction (MIP) Model}
\label{subsec:mip_model}

\begin{figure}[htb]
\centering
\includegraphics[width=0.9\textwidth]{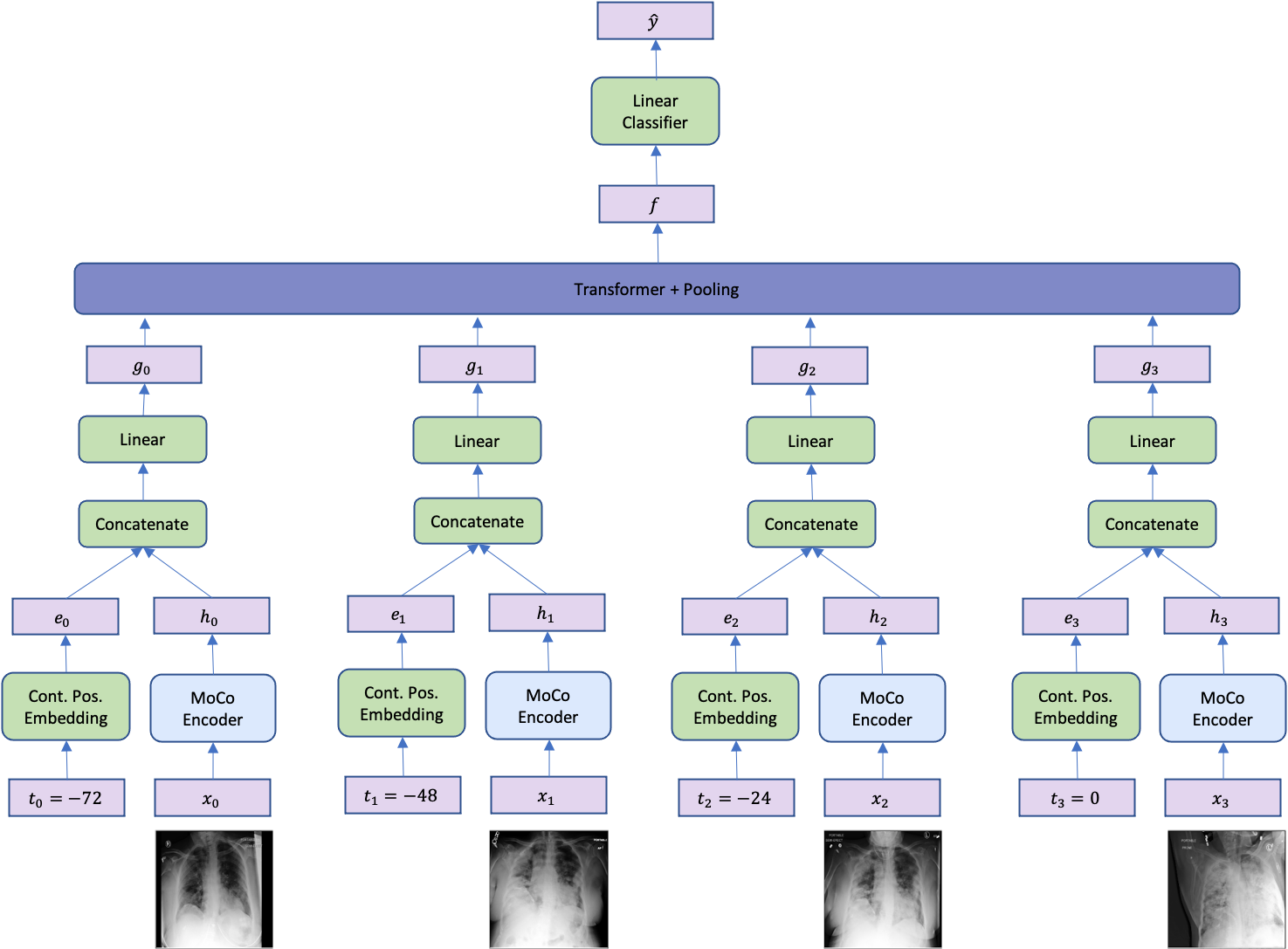}
\caption{Schematic of the Multiple Image Prediction (MIP) model. The MIP model takes a sequence of images along with their relative scan times as input and outputs a binary prediction for each adverse event. The MoCo encoder model is applied to each image to learn image representations. These image representations are concatenated with a time embedding obtained from the relative scan times and then projected to a lower dimension and then input to a transformer network to aggregate information from all images. The output of the transformer is pooled together and input to a linear classifier.}
\label{fig:transformer}
\end{figure}

For the MIP task, we propose a new model that takes a sequence of X-ray images $(x_0, ..., x_n)$ along with their scan times $(t_0, ..., t_n)$ relative to the final scan as inputs and predicts the likelihood of adverse events occurring after the final scan.
The overall model structure is shown in Figure \ref{fig:transformer}.
The scan times are represented as the number of hours from the final scan time. Thus, if a patient had two previous scans done, e.g., 50 hours and 20 hours before the final scan, then the scan times will be represented as $(t_0 = -50, t_1 = -20, t_2 = 0)$. The final scan time $t_n$ is always 0.

During the forward pass, each image $x_i$ is first passed to the MoCo encoder model in parallel to obtain an image representation $h_i$. Separately, each relative scan time $t_i$ is passed to a \emph{Continuous Position Embedding (CPE)} module to learn a time embedding $e_i$. The CPE module, described in more detail below, maps each time point to a different embedding. The two representations $h_i$ and $e_i$ for each $i$ are then concatenated together and then projected to a lower dimension using a fully connected layer. The full sequence of images is then input to a transformer network \cite{vaswani2017attention}. The transformer network maps an input sequence to an output sequence of the same length using a self-attention mechanism. The output of the transformer is the sum-pooled to obtain a hidden representation $f$ that contains aggregated information from all images. This hidden representation $f$ is input to a linear classifier.

Similar to the SIP task, the weights of the MoCo encoder are initialized to the pretrained weight values while the weights of the transformer and the linear classifier are randomly initialized. The entire network is fine-tuned jointly on a labeled dataset to mimize the binary cross entropy loss.

\paragraph{Continuous Positional Embedding (CPE)} The CPE, inspired by the positional embedding (PE) from \cite{vaswani2017attention}, is designed to map each time point to a $d$-dimensional vector representation. The CPE has the same functional form as the PE, but it can take continuous values within a certain range as input. Formally, the CPE maps a relative scan time $s$ to a vector $e$ as:

\begin{equation}
\begin{split}
    e_{(t, 2i)} &= \sin \left ( t / 10000^{2i / d} \right ),\\
    e_{(t, 2i+1)} &= \cos \left ( t / 10000^{2i / d} \right ),
\end{split}
\end{equation}

where $t \in [0, 360)$. We only include X-ray scans taken within the last 360 hours (15 days) of the end exam since the COVID-19 infection is unlikely to last longer than 15 days.

\paragraph{DropImage Regularizer} To reduce overfitting and to improve the robustness of the MIP model, we apply a masking based regularizer that we call the \emph{DropImage} regularizer. The DropImage regularizer, when applied to a an image sequence, drops a subset of images chosen independently at random. The final image is never removed. DropImage encourages learning more robust models that can make good predictions even if some of the past images were missing. Further, DropImage can also be viewed as a data augmentation scheme.

\section{Experiments}

\subsection{Pretraining}

We applied both supervised and self-supervised pretraining procedures.
Our supervised pretraining was similar to that for CheXpert \cite{irvin2019chexpert}.
For supervised pretraining, we trained the DenseNet-121 model with the Adam optimizer \cite{kingma2015adam} with a learning rate of $10^{-3}$, weight decay of $10^{-5}$, and batch size of 64 on the MIMIC-JPG dataset \cite{johnson2019mimicjpg}.
Data augmentation included interpolation to a 224 $\times$ 224 grid and random vertical/horizontal flipping.
We pretrained for 10 epochs, decaying the learning rate by a factor of 10 each epoch.

We pretrained our self-supervised model using the momentum contrast training described in Section \ref{subsec:selfsup_pretraining}.
For data augmentation, we used random cropping and interpolation to a 224 $\times$ 224 grid, random horizontal/vertical flipping, and random Gaussian noise addition.
We investigated a number of augmentation strategies for pretraining the MoCo models, including noise additions, affine transformations, color transformations, and X-ray acquisition simulation, but in the end we found that these provided little benefit beyond those in the original MoCo paper \cite{he2020momentum}.
After the augmentations, we applied histogram normalization.
The histogram normalization procedure preserves the relative values of the pixels by interpolating along a shifted version of the image pixel value histogram while constraining the resulting pixel values to be in the specified target range.
The augmentations are described in further detail in Appendix \ref{app:data_aug}.

We tuned the following hyperparameters during the pretraining phase: learning rate, MoCo latent feature dimension size, and the queue size.
We searched over a logarithmic scale of values, varying the learning rate within $10^{\{-2, -1, 0\}}$, and MoCo feature dimensions within $\{64, 128, 256\}$.
The queue size was fixed at 65,536.
We used a batch size of 128 for each of 8 GPUs, the largest we could achieve in initial testing, accumulating gradients using PyTorch's DistributedDataParallel framework \cite{li2020pytorch,he2020momentum}.
We selected hyperparameters based a cross-validation analysis on the downstream tasks.
We optimized models using stochastic gradient descent with momentum \cite{sutskever2013importance}, using 0.9 as the momentum term and a weight decay parameter of $10^{-4}$.
Pretraining took approximately four days using eight 16 GB Nvidia V100 GPUs.

\subsection{Example Downstream Task Outputs}

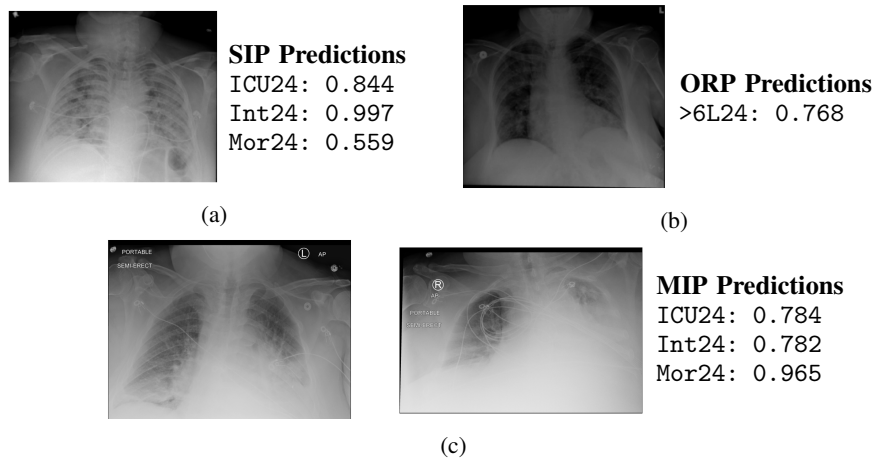
\begin{figure}[htb]
    \centering
    \begin{subfigure}{.35\textwidth}
    \centering
        \begin{tikzpicture}
            \node[inner sep=0pt] (base_image) at (0,0)
                {\includegraphics[width=.5\textwidth,]
                {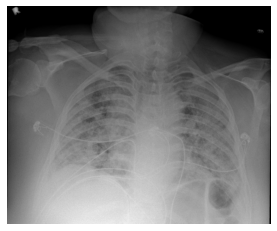}};
            \node[fill=white,text width=2.5cm] (sip_outputs) at (2.8,0)
                {\textbf{SIP Predictions} \\
                \texttt{ICU24: 0.844} \\
                \texttt{Int24: 0.997} \\
                \texttt{Mor24: 0.559}};
        \end{tikzpicture}
        \caption{}
        \label{subfig:sip_predictions}
    \end{subfigure}
    \begin{subfigure}{.38\textwidth}
    \centering
        \begin{tikzpicture}
            \node[inner sep=0pt] (base_image) at (0,0)
                {\includegraphics[width=.45\textwidth,]
                {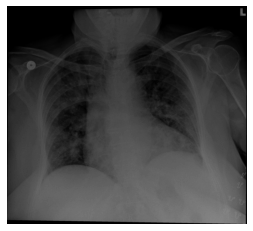}};
            \node[fill=white,text width=2.7cm] (orp_outputs) at (2.9,0)
                {\textbf{ORP Predictions} \\
                \texttt{>6L24: 0.768}};
        \end{tikzpicture}
        \caption{}
        \label{subfig:orp_predictions}
    \end{subfigure}
    
    \begin{subfigure}{.67\textwidth}
    \centering
        \begin{tikzpicture}
            \node[inner sep=0pt] (base_image) at (0,0)
                {\includegraphics[trim=80 80 80 80,clip,width=.75\textwidth,]
                {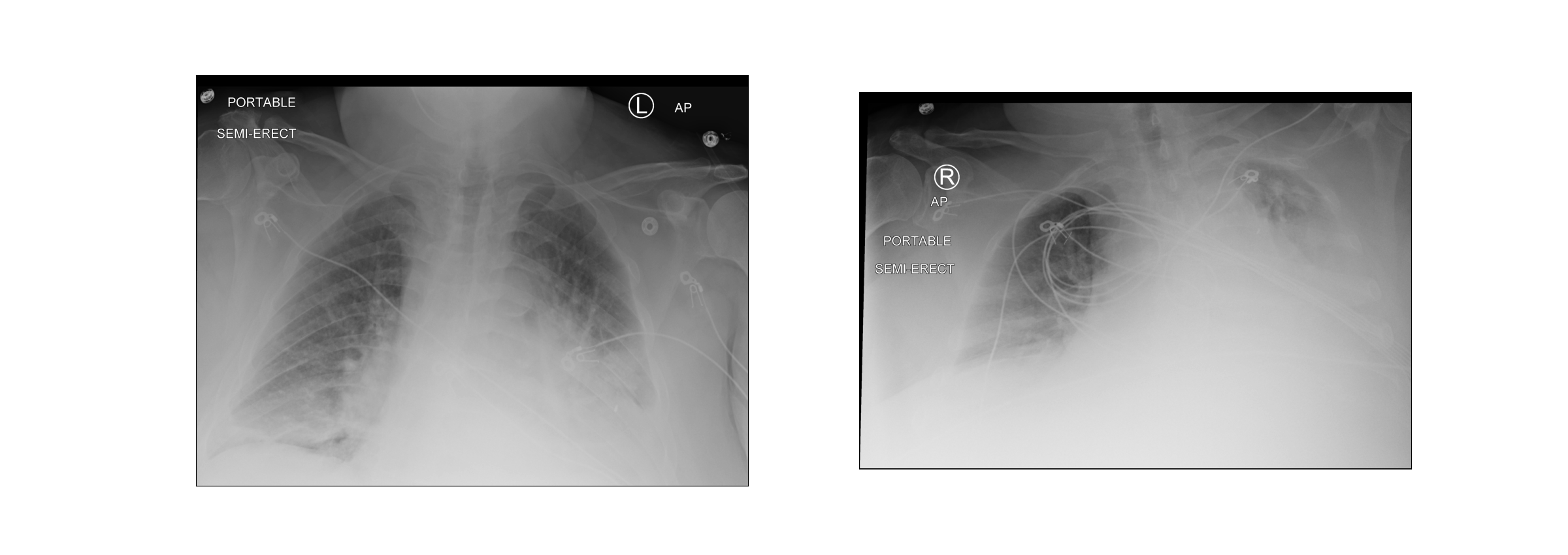}};
            \node[fill=white,text width=2.5cm] (mip_outputs) at (5.1,0)
                {\textbf{MIP Predictions} \\
                \texttt{ICU24: 0.784} \\
                \texttt{Int24: 0.782} \\
                \texttt{Mor24: 0.965}};
        \end{tikzpicture}
        \caption{}
        \label{subfig:mip_predictions}
    \end{subfigure}
    \caption{Example model outputs for ICU (\texttt{ICU24}), intubation (\texttt{Int24}) mortality (\texttt{Mor24}), and oxygen greater than 6 L per day (\texttt{>6L24}) prediction tasks, all at 24 hours. (\protect \subref{subfig:sip_predictions}) Example SIP outputs based on a single image with evident increased lung opacity. In this case the patient did not suffer any adverse event in the next 24 hours, but would ultimately suffer all three adverse events within 72 hours. (\protect \subref{subfig:orp_predictions}) Example ORP output based on a single image. This patient required greater than 6 L per day of oxygen within 24 hours. (\protect \subref{subfig:mip_predictions}) Example MIP outputs. Both images were taken from the same patient with 49 hours of separation. Increased lung opacity is observable in the second (later) image. The patient suffered all three adverse events within 24 hours.}
    \label{fig:model_examples}
\end{figure}

Figure \ref{fig:model_examples} illustrates the frameworks for each of the prediction tasks over the succeeding 24 hours.
Figure \ref{subfig:sip_predictions} shows an example of SIP predictions from a patient with increased lung opacity.
In this case the patient did not suffer any adverse event in the next 24 hours, but ultimately suffered all three adverse events within 72 hours.
Figure \ref{subfig:orp_predictions} shows an example of ORP predictions from a patient that required increased oxygeen within 24 hours.
In Figure \ref{subfig:mip_predictions}, a sequence of chest X-rays with increasing lung opacity is used by the MIP model to predict COVID deterioration for a patient.
The images were taken 49 hours apart.
For the case of Figure \ref{subfig:mip_predictions}, the patient was transferred to the ICU, intubated, and suffered a mortality within 24 hours.

\subsection{Single-Image Task Ablations}
\label{subsec:delong_intervals}

We considered both self-supervised and supervised pretraining for each of our single-image tasks (SIP and ORP).
Data splits for these tasks are shown in Table \ref{tab:dataset_summary}.
Following previous work \cite{irvin2019chexpert,zhang2020diagnosis,Kwon2020CombiningIR}, we compare models via their areas under receiver operator characteristic curve (AUC).
In order to facilitate a straightforward comparisons, our pretraining in this section for both supervised and MoCo only used the MIMIC dataset.

Before fine-tuning for the SIP and ORP tasks, the weights of the encoder are initialized to the pretrained weight values while the weights of the linear classifier are randomly reinitialized with a uniform distribution \cite{he2015delving}.
We considered three different fine-tuning ablations.
The first ablation (CL) fixed the MoCo encoder (including BatchNorm statistics) and only trained the new linear classifier \cite{he2020momentum}.
The second ablation (FT) allowed the entire model to train, not just the classifier.
The final ablation (FT RA) allowed the entire model to train and also incorporated further random augmentations (rotation, X-shear, Y-shear, and translations).
The CL model was fine-tuned for 5 epochs. The FT models were fine-tuned for 20 epochs, and the FT RA models were fine-tuned for 40 epochs.
We also show a ``Scratch'' ablation, which is a DenseNet model with random initialization \cite{he2015delving} trained only on the COVID data (i.e., no pretraining).

We tuned each ablation using a stratified 5-fold cross validation of the \{train,val\} split of the data.
We tuned learning rate in the range $10^{\{-4, -3, -2, -1\}}$ and optimized over all MoCo/supervised-pretrained models for model selection.
We also tuned whether to use the Adam \cite{kingma2015adam} or SGD \cite{sutskever2013importance} optimizers.
All fine-tunings used the cosine annealing learning rate decay \cite{Loshchilov2017SGDRSG}.
After the cross-validation, we selected the best model for each ablation and applied a bootstrap analysis to estimate performance characteristics on the test set.

To test for significance between methods we applied 1,000 bootstrap iterations to the test set and computed the difference in AUC between the best MoCo method and the best supervised pretraining method.
Methods where the best MoCo AUC is significantly higher than the best supervised AUC are indicated with *.
Our tables show results in two rows for each method.
The top row shows the holdout test AUC, while the bottom row shows percentile bootstrap 95\% confidence intervals.
In some cases we have statistical significance with overlapping confidence intervals--this is because the two methods are correlated, an aspect captured in the bootstrapped AUC differences.
We chose this non-parametric test over the standard method of DeLong \cite{delong1988comparing,sun2014fast} to maintain a consistent presentation, but in a separate analysis found the DeLong AUC difference test generally matched our non-parametric test.

\begin{table}[htb]
    \caption{SIP AUCs Across Ablations (95\% CI: Bootstrap)}
    \centering
    \begin{tabular}{lcccc}
        \toprule
        & \multicolumn{4}{c}{AUC of Any Adverse Event Prediction} \\
        \cmidrule{2-5}
        & 24 hours & 48 hours & 72 hours & 96 hours \\
        \midrule
        \multirow{2}{*}{Scratch FT (No PT)} & 0.616 & 0.645 & 0.651 & 0.667 \\
        & (0.550, 0.679) & (0.588, 0.701) & (0.598, 0.705) & (0.620, 0.715) \\
        \midrule
        \multirow{2}{*}{Supervised PT CL} & 0.668 & 0.690 & 0.681 & 0.692 \\
        & (0.599, 0.738) & (0.625, 0.753) & (0.628, 0.730) & (0.645, 0.743) \\
        \multirow{2}{*}{Supervised PT FT} & 0.672 & 0.697 & 0.687 & 0.702 \\
        & (0.610, 0.739) & (0.639, 0.754) & (0.631, 0.737) & (0.653, 0.750) \\
        \multirow{2}{*}{Supervised PT FT RA} & 0.658 & 0.694 & 0.686 & 0.703 \\
        & (0.591, 0.722) & (0.638, 0.747) & (0.635, 0.736) & (0.658, 0.748) \\
        \midrule
        \multirow{2}{*}{Moco PT CL} & \textbf{0.691} & 0.716 & 0.711 & 0.730 \\
        & (0.627, 0.748) & (0.662, 0.766) & (0.660, 0.758) & (0.686, 0.775) \\
        \multirow{2}{*}{Moco PT FT} & 0.686 & \textbf{0.719} & \textbf{0.726*} & \textbf{0.742*} \\
        & (0.619, 0.751) & (0.659, 0.769) & (0.677, 0.774) & (0.696, 0.785) \\
        \multirow{2}{*}{Moco PT FT RA} & 0.658 & 0.703 & 0.708 & 0.729 \\
        & (0.590, 0.724) & (0.646, 0.761) & (0.656, 0.758) & (0.683, 0.772) \\
        \bottomrule
    \end{tabular}
    \label{tab:covid_sip_results_delong}
\end{table}
A comparison via AUC values for supervised vs. self-supervised pretraining for the SIP task is shown in Table \ref{tab:covid_sip_results_delong}.
AUC values generally increased over longer time intervals.
MoCo models had the highest AUC scores for predicting adverse events at all time windows, achieving significance at 72 and 96 hours.
The MoCo PT CL model (pretrained with MoCo, only fine-tuning classification layer) had the highest AUC at 24 hours, whereas the PT FT model (which allowed the entire model to fine-tune) achieved highest AUC at 48, 72 and 96 hours.
The Scratch method performed worst across all architectures and pretraining methods.
Performance improved between Scratch and supervised pretraining, and then again between supervised pretraining and MoCo pretraining.

\begin{table}[htb]
    \caption{ORP AUCs Across Ablations (95\% CI Method: Bootstrap)}
    \centering
    \begin{tabular}{lcccc}
        \toprule
        & \multicolumn{4}{c}{AUC of O2 $>$ 6L requirement Prediction} \\ \cmidrule{2-5}
        & 24 hours & 48 hours & 72 hours & 96 hours \\
        \midrule
        \multirow{2}{*}{Scratch FT (No PT)} & 0.696 & 0.670 & 0.654 & 0.632 \\
        & (0.636, 0.754) & (0.619, 0.720) & (0.611, 0.697) & (0.591, 0.674) \\ \midrule
        \multirow{2}{*}{Supervised PT CL} & 0.749 & 0.706 & \textbf{0.712} & \textbf{0.706} \\
        & (0.698, 0.793) & (0.659, 0.750) & (0.672, 0.751) & (0.671, 0.741) \\
        \multirow{2}{*}{Supervised PT FT} & 0.727 & 0.701 & 0.707 & 0.694 \\
        & (0.676, 0.777) & (0.655, 0.746) & (0.668, 0.746) & (0.658, 0.729) \\
        \multirow{2}{*}{Supervised PT FT RA} & 0.715 & 0.685 & 0.695 & 0.679 \\
        & (0.655, 0.769) & (0.641, 0.731) & (0.657, 0.731) & (0.644, 0.716) \\ \midrule
        \multirow{2}{*}{Moco PT CL} & 0.699 & 0.683 & 0.690 & 0.686 \\
        & (0.644, 0.750) & (0.635, 0.724) & (0.654, 0.728) & (0.652, 0.722) \\
        \multirow{2}{*}{Moco PT FT} & \textbf{0.765*} & \textbf{0.713} & 0.702 & 0.688 \\
        & (0.719, 0.807) & (0.671, 0.753) & (0.666, 0.737) & (0.654, 0.722) \\
        \multirow{2}{*}{Moco PT FT RA} & 0.722 & 0.696 & 0.698 & 0.694 \\
        & (0.671, 0.767) & (0.651, 0.739) & (0.661, 0.734) & (0.658, 0.725) \\
        \bottomrule
    \end{tabular}
    \label{tab:o2_results_delong}
\end{table}
A comparison via AUC values for Supervised vs. Self-Supervised pretraining for the ORP task is shown in Table~\ref{tab:o2_results_delong}.
The time course trend is inverted relative to Table~\ref{tab:covid_sip_results_delong}, with O2 models becoming less accurate at larger time windows.
In this case, Supervised pretraining models performed more similarly to MoCo-pretrained models.
The MoCo PT FT model performing had the best AUC at 24 and 48 hours, while the Supervised PT CL model had the best AUC at 72 and 96 hours.
Only 24 hour prediction using MoCo was significant based on the bootstrap tests.
% However, none of the differences were significant with the bootstrap tests.

\subsection{SIP Comparison to COVID-GMIC}
\label{subsec:bootstrap_exp}

COVID-GMIC is a previous neural network image analysis model developed for the purpose of COVID prognosis prediction \cite{shamout2020artificial}.
COVID-GMIC utilizes supervised pretraining with an architecture explicitly designed for synthesizing information at coarse and fine image scales \cite{shen2020interpretable} with supervised pretraining on the NIH dataset \cite{wang2017chestx}.
Our SIP experiments used the same test set that was used for the COVID-GMIC, so we compare our models directly in Table \ref{tab:gmic_comp}.
For this comparison we included a different MoCo pretraining procedure where we used a combination of both the MIMIC-CXR \cite{johnson2019mimic} and CheXpert \cite{irvin2019chexpert} data sets for pretraining.
Despite having more data, in our 5-fold cross-validation of the {train,val} split the same the same model was selected as in Section \ref{subsec:delong_intervals}.
\begin{table}[htb]
    \centering
    \caption{SIP AUCs Across Methods (95\% CI Method: Bootstrap)}
    \begin{tabular}{lcccc}
        \toprule
        & \multicolumn{4}{c}{AUC of Any Adverse Event Prediction} \\ \cmidrule{2-5} 
        & 24 hours & 48 hours & 72 hours & 96 hours \\ \midrule
        % MoCo Pre-training best
        \multirow{2}{*}{DenseNet MoCo (best)} & 0.686 & \textbf{0.719} & \textbf{0.726} & \textbf{0.742} \\
        & (0.619, 0.751) & (0.659, 0.769) & (0.677, 0.774) & (0.696, 0.785) \\
        % GMIC
        \multirow{2}{*}{COVID-GMIC} & \textbf{0.695} & 0.716 & 0.717 & 0.738 \\
        & (0.627, 0.754) & (0.661, 0.766) & (0.661, 0.766) & (0.691, 0.781) \\ \bottomrule
    \end{tabular}
    \label{tab:gmic_comp}
\end{table}

In this case we did not apply significance testing since we did not have the raw predictions from this paper.
Both the MoCo and COVID-GMIC methods are comparable across time points, with wide overlaps in confidence intervals.
The widest gap in performance was for predicting adverse events in the next 24 hours, where COVID-GMIC had a 0.02 higher AUC.
At longer time intervals the performance of MoCo methods began to improve, showing higher AUC scores than COVID-GMIC on the test set at 48, 72 and 96 hours.

\subsection{Multi-Image Task Ablations}

Table \ref{tab:mip_ablations} shows ablations for the MIP task.
In this case, we compared and tested the difference between the MoCo PT CL ablation and the Transformer model built on top of MoCo representations. We trained each model with the Adam optimizer \cite{kingma2015adam} for 50 epochs with a batch size of 32. We tuned the following hyperparameters using grid search: learning rate, drop image probability ($p_{drop}$), projection dimension and the pooling method. We searched over learning rate values of $10^{\{-3,-2,-1\}}$, $p_{drop}$ values in $\{0, 0.1, 0.2, 0.5\}$, projection dimension in $\{16, 32, 64, 128\}$. For the pooling method, we used either sum pooling or simply taking the final time step. We set the dropout value for the transformer and the CPE to be $0.5$ and weight decay to $10^{-5}$.
\begin{table}[htb]
    \centering
    \caption{MIP AUCs Across Ablations (95\% CI Method: Bootstrap)}
    \begin{tabular}{lcccc}
        \toprule
        & \multicolumn{4}{c}{AUC of Any Adverse Event Prediction} \\ \cmidrule{2-5} 
        & 24 hours & 48 hours & 72 hours & 96 hours \\
        \midrule
        \multirow{2}{*}{MoCo PT CL} & 0.747 & 0.755 & 0.758 & 0.768 \\
        & (0.719, 0.776) & (0.730, 0.780) & (0.734, 0.782) & (0.746, 0.790) \\
        % \multirow{2}{*}{Supervised PT + Transformer} & 0.739 & 0.746 & 0.744 & 0.745 \\
        % & (0.711, 0.768) & (0.720, 0.770) & (0.721, 0.767) & (0.722, 0.770) \\
        \multirow{2}{*}{MoCo PT + Transformer} & \textbf{0.767*} & \textbf{0.778*} & \textbf{0.778*} & \textbf{0.786} \\
        & (0.740, 0.796) & (0.752, 0.802) & (0.755, 0.800) & (0.764, 0.807) \\
        \bottomrule
    \end{tabular}
    \label{tab:mip_ablations}
\end{table}

The Transformer method performed better than the single-image MoCo method for predicting adverse events at every time window with statistical significance at 24, 48, and 72 hours.
A more detailed version of Table \ref{tab:mip_ablations} comparing Single-image and multi-image prediction for other adverse events is shown in Table\ref{tab:mip_bootstrap} of Appendix \ref{app:bootstrap}.
The MIP Transformer model was also better at predicting ICU transfers, intubations, and mortalities.
The improvements for ICU transfers were significant at all time windows, while mortalities were significant at 72 and 96 hours.
The performance differences for intubation were not statistically significant.

\subsection{Reader Study}
We compared the performance of our model with two chest radiologists from NYU Langone Health in a pilot reader study with a sample of 200 image sequences from the test set (note: this was a new, distinct reader study from that used previously \cite{shamout2020artificial}).
The results of the study for ``any'' adverse events are shown in Table \ref{tab:reader_study_small}, while Table \ref{tab:reader_study_extended} in Appendix \ref{app:reader_extended} shows an extended version of the data with breakdowns by adverse event category.
We also present the results from averaging the prediction of both radiologists (shown as ``Radiologist A + B'').
\begin{table}[htb]
\centering
% \resizebox{\textwidth}{!}{%
\caption{MIP AUCs Compared to Radiologists (95\% CI Method: Bootstrap)}
\begin{tabular}{lcccc}
\toprule
 & \multicolumn{4}{c}{AUC of Any Adverse Event Prediction} \\ \cmidrule{2-5} 
 & 24 hours & 48 hours & 72 hours & 96 hours \\ \midrule
\multirow{2}{*}{MoCo PT + Transformer} & 0.785 & \textbf{0.801} & \textbf{0.790} & \textbf{0.790} \\
            & (0.717, 0.852) & (0.739, 0.861) & (0.726, 0.855) & (0.727, 0.853) \\
\multirow{2}{*}{Radiologist A} &  0.786 & 0.769 & 0.733 & 0.739 \\
            & (0.694, 0.840) & (0.709, 0.840) & (0.664, 0.804) & (0.658, 0.797) \\
\multirow{2}{*}{Radiologist B} &  \textbf{0.789} & 0.787 & 0.762 & 0.751 \\
            & (0.709, 0.846) & (0.716, 0.846) & (0.693, 0.826) & (0.684, 0.816) \\
\multirow{2}{*}{Radiologist A + B} &  0.784 & 0.787 & 0.761 & 0.754 \\
            & (0.703, 0.774) & (0.722, 0.786) & (0.687, 0.754) & (0.678, 0.745) \\
 \bottomrule
 \end{tabular}
\label{tab:reader_study_small}
\end{table}

For this task the radiologists indicated the presence of intubation support devices in the X-ray.
We performed an audit of our selection criteria and found that in some cases these patients did not have an intubation event in the anonymized electronic health records despite having the support devices visible on the chest X-ray.
In other cases, the adverse event was recorded later.
Given that images with support devices were likely in the training set, we allowed radiologists to simply predict positive for any cases where they saw a support device, as this information was available to the model as well.

Both radiologists exhibited wide confidence intervals as indicated by bootstrap methods.
Radiologist AUC values trended better than our model for the task of intubation prediction (Appendix \ref{app:reader_extended}), while our model trended better for ICU transfer and mortality prediction.
On the task of predicting any adverse event, our model performed better for longer term prediction.
The largest differences for AUC scores between radiologists and our model occurred when predicting mortalities, where our model had better AUC scores.
Radiologists were aware of the primary timeframe of the dataset (i.e., April-June 2020).

\section{Discussion}

\subsection{Performance Among Proposed Models}

Our study examined whether performance for COVID prognosis prediction could be improved via the use of two techniques: 1) self-supervised pretraining and 2) prediction based on image sequences.
Self-supervised pretraining achieved the highest AUC scores on our hold-out test set for all SIP tasks of interest.
Our bootstrap statistical tests found significance at 72 and 96 hours.
For ORP tasks, supervised pretraining performed similarly to MoCo pretraining, except for 24 hours.
This may indicate that features associated with pathological lung findings are more predictive for oxygen requirements than for adverse events, but this topic needs to be studied further.

Interestingly, the predictions for the SIP tasks became more accurate over longer time intervals, whereas the predictions for the ORP tasks became more accurate over shorter time intervals.
The SIP results could have improved due to the presence of more positive labels at the later time windows.
Label scarcity hindered model performance in all test sets, as indicated by the bootstrap confidence intervals in Section \ref{subsec:bootstrap_exp} and Appendix \ref{app:bootstrap}.
However, the inversion of the trend for ORP cannot be explained by label scarcity.
It could be that there are image features that become readily apparent for the ORP task close to the actual moment of increased oxygen needs.

Our greatest performance gain occurred when when predicting based on image sequences rather than single images.
These improvements were statistically significant via our non-parametric tests.
Radiologist sentiment indicated that it is difficult to identify patient conditions from single images, as temporal trajectories are critical for performing analyses.
The results of our transformer models support this intuition, with AUC scores improving for all tasks by at least 0.01 and as much as 0.05 when using image sequences rather than individual images.

\subsection{Test Set Uncertainty}

A recent meta-study noted that many prediction models targeting COVID are at a high risk of overfitting and bias \cite{wynants2020prediction}.
We addressed this via two methods: 1) pretraining on non-COVID data and 2) explicit sample selection investigation via bootstrap experiments.
We observed that the final bootstrap confidence intervals in Section \ref{subsec:bootstrap_exp} were wide enough to include broad classes of ablations, corroborating the uncertainties associated with small datasets.
The greatest sensitivity was with respect to positive labels, which were infrequent among the test splits for the various tasks.
It would be reasonable to expect forward performance anywhere within the range of the bootstrapped confidence intervals.
This uncertainty is on top of other caveats already well-known in the medical field, namely, uncertainty arising due to changes in the patient population and treatment effects.

\subsection{Pilot Clinical Study Limitations}

Our pilot clinical study found that our multi-image prediction model had comparable performance to human radiologists for most tasks and surpassed human radiologists in predicting mortalities.
However, this study has some limitations.
We asked radiologists to rate the probability of adverse events over the upcoming time windows.
Depending on the hospital and local practices, this may not be a part of the standard radiologist workflow.
Radiologists are typically asked to summarize imaging findings and to convey nuance with how the information in image features may impact clinical care.
It is possible that with further calibration, radiologist performance could be improved for the task of adverse event prediction.
We present these results to help readers put model performance in the context of currently possible clinical predictions.

\subsection{Alternative Approaches}

Supervised DenseNet pretraining was previously used \cite{shamout2020artificial} to generate higher AUC scores than we report here.
The previous work included further optimizations, including a larger hyperparameter sweep and averaging predictions of several models.
Some of these optimizations - in particular model ensembling - were not investigated in our study, but could lead to a closing of the gap in performance between supervised and self-supervised models that we observe in this paper.

Since the start of our work, new approaches for self-supervision have been published in the literature.
Clustering can be used to relax the approach in this paper that only looks at a single image for matching \cite{caron2020unsupervised}.
Based on our ORP results, another approach that might be beneficial would be including labels as a form of weak supervision along side a constrastive loss.
Investigating these alternative approaches may provide further performance improvements for the self-supervised methods.

\section{Conclusion}
The COVID-19 pandemic has created a major need for risk stratification of patients in clinical settings.
Developing AI methods for this disease is difficult, as acquisition of large datasets is difficult or impossible for most medical centers.
In this paper we attempted to address this problem via self-supervised contrastive loss pretraining.
We found that our single-image prediction model was able to match a previous model \cite{shamout2020artificial} using separate advances, opening up new possibilities for research on synergies between contrastive pretraining and model architecture design.
Our multi-image model performance surpassed that of all single-image models. In comparison to radiologists, our multi-image prediction model was comparable in its ability to predict patient deterioration and stronger in its ability to predict mortality.
We hope these contributions will assist the community going forward on the task of hospital resource planning.

\section{Acknowledgments}
We would like to thank the IT team at the NYU hospital for their help in acquiring and curating COVID-19 X-ray data.
We thank the members of the NYU medical vision team for helpful discussions and insights on X-ray data.
We thank Jean-Remi King, C. Lawrence Zitnick, Mark Tygert, Marcus Rohrbach, Tullie Murrell, Michal Drozdzal, Adriana Romero and Lucas Caccia from Facebook AI Research for useful conversations on technical components of the work.
We also thank Michael Chasse and Louis Mullie from University of Montreal and CHUM for sharing their knowledge on interpretation of COVID patient data.
We would like to thank the Stanford CheXpert team for providing their dataset for this research.
Additionally, this work at NYU was supported in part by a grant from the National Institutes of Health (P41EB017183).

\appendix
\setcounter{table}{0}
\renewcommand{\thetable}{A\arabic{table}}
\section{Pretraining Augmentation Parameters}
\label{app:data_aug}
For self-supervised momentum contrast pretraining, we used the following augmentations (in order): random resizing/cropping, random horizontal flipping, random vertical flipping, random Gaussian blur, Gaussian noise addition, and histogram normalization.
For each sample, each random augmentation was applied with probability $p=0.5$.
The cropping from the random resizing/cropping augmentation was done at an image scale uniformly distributed between 20\% and 100\% the size of the original image.

For the blur augmentation, we applied the following normalized Gaussian kernel:
\begin{equation}
    g(x,y) = \frac{1}{\sigma_{\text{kernel}} \sqrt{2\pi}} \text{exp} \left ( -\frac{1}{2} \frac{x^2 + y^2}{\sigma_{\text{kernel}}^2} \right ),
\end{equation}
where $\sigma$ was selected for each sample uniformly at random between 0.1 and 2.0 pixels.

We selected the standard deviation for the noise addition randomly according to the following formula:
\begin{equation}
    \sigma_{\text{noise}} = \frac{\mu_{\text{image}}}{\text{SNR}},
\end{equation}
where $\text{SNR}$ was selected uniformly between 4 and 8 for each sample and $\mu_{\text{image}}$ was the average pixel value of the input sample image.

\section{Bootstrap Results}
\label{app:bootstrap}

\begin{table}[htb]
    \centering
    \caption{MIP AUCs Across Methods (95\% CI Method: Bootstrap)}
    \begin{tabular}{lcccc}
        \toprule
        & \multicolumn{4}{c}{AUC of Any Adverse Event Prediction} \\ \cmidrule{2-5} 
        & 24 hours & 48 hours & 72 hours & 96 hours \\
        \midrule
        \multirow{2}{*}{DenseNet MoCo PT} & 0.747 & 0.755 & 0.758 & 0.768 \\
        & (0.719, 0.776) & (0.730, 0.780) & (0.734, 0.782) & (0.746, 0.790) \\
        \multirow{2}{*}{MoCo PT + Transformer} & \textbf{0.767*} & \textbf{0.778*} & \textbf{0.778*} & \textbf{0.786} \\
        & (0.740, 0.796) & (0.752, 0.802) & (0.755, 0.800) & (0.764, 0.807) \\
        \midrule
        
        & \multicolumn{4}{c}{AUC of ICU Prediction} \\ \cmidrule{2-5} 
        & 24 hours & 48 hours & 72 hours & 96 hours \\ \midrule
        \multirow{2}{*}{DenseNet MoCo PT} & 0.705 & 0.706 & 0.706 & 0.712 \\
        & (0.673, 0.747) & (0.674, 0.740) & (0.672, 0.734) & (0.679, 0.738) \\
        \multirow{2}{*}{DenseNet MoCo PT + Transformer} & \textbf{0.725*} & \textbf{0.740*} & \textbf{0.738*} & \textbf{0.746*} \\
        & (0.672, 0.788) & (0.686, 0.798) & (0.689, 0.794) & (0.697, 0.789) \\
        \midrule
        
        & \multicolumn{4}{c}{AUC of Intubation Prediction} \\ \cmidrule{2-5} 
        & 24 hours & 48 hours & 72 hours & 96 hours \\ \midrule
        \multirow{2}{*}{DenseNet MoCo PT} & 0.719 & 0.710 & 0.705 & \textbf{0.700} \\
        & (0.681, 0.757) & (0.674, 0.745) & (0.672, 0.737) & (0.668, 0.732) \\
        \multirow{2}{*}{DenseNet MoCo PT + Transformer} & \textbf{0.794} & \textbf{0.754} & \textbf{0.725} & 0.690 \\
        & (0.721, 0.863) & (0.685, 0.822) & (0.646, 0.791) & (0.635, 0.751) \\
        \midrule
        
        & \multicolumn{4}{c}{AUC of Mortality Prediction} \\
        \cmidrule{2-5} 
        & 24 hours & 48 hours & 72 hours & 96 hours \\ \midrule
        \multirow{2}{*}{DenseNet MoCo PT} & 0.812 & 0.822 & 0.822 & 0.817 \\
        & (0.769, 0.855) & (0.790, 0.854) & (0.792, 0.852) & (0.792, 0.843)  \\
        \multirow{2}{*}{DenseNet MoCo PT + Transformer} & \textbf{0.825} & \textbf{0.827} & \textbf{0.825*} & \textbf{0.848*} \\
        & (0.777, 0.870) & (0.796, 0.860) & (0.795, 0.855) & (0.819, 0.865) \\
        \bottomrule
    \end{tabular}
    \label{tab:mip_bootstrap}
\end{table}
The extended results for comparing single-image prediction to the MIP method are shown in Table \ref{tab:mip_bootstrap} (* indicates bootstrap significance at $\alpha=0.05$).
In this case, the Transformer-based MIP model gave larger advantages over the SIP MoCo pretrained models, with some labels being significant.
This indicates the benefits when incorporating image sequences for COVID prognosis prediction.

\section{Reader Study (Extended)}
\label{app:reader_extended}
\begin{table}[htb]
\centering
\caption{MIP AUCs Compared to Radiologists (95\% CI Method: Bootstrap)}
\begin{tabular}{lcccc}
\toprule
 & \multicolumn{4}{c}{AUC of Any Adverse Event Prediction} \\ \cmidrule{2-5} 
 & 24 hours & 48 hours & 72 hours & 96 hours \\ \midrule
\multirow{2}{*}{DenseNet MoCo PT + Transformer} & 0.785 & \textbf{0.801} & \textbf{0.790} & \textbf{0.790} \\
            & (0.717, 0.852) & (0.739, 0.861) & (0.726, 0.855) & (0.727, 0.853) \\
\multirow{2}{*}{Radiologist A} &  0.786 & 0.769 & 0.733 & 0.739 \\
            & (0.694, 0.840) & (0.709, 0.840) & (0.664, 0.804) & (0.658, 0.797) \\
\multirow{2}{*}{Radiologist B} &  \textbf{0.789} & 0.787 & 0.762 & 0.751 \\
            & (0.709, 0.846) & (0.716, 0.846) & (0.693, 0.826) & (0.684, 0.816) \\
\multirow{2}{*}{Radiologist A + B} &  0.784 & 0.787 & 0.761 & 0.754 \\
            & (0.703, 0.774) & (0.722, 0.786) & (0.687, 0.754) & (0.678, 0.745) \\
 \midrule

 & \multicolumn{4}{c}{AUC of ICU Prediction} \\ \cmidrule{2-5} 
 & 24 hours & 48 hours & 72 hours & 96 hours \\ \midrule
\multirow{2}{*}{DenseNet MoCo PT + Transformer} & \textbf{0.773} & \textbf{0.774} & \textbf{0.751} & \textbf{0.747} \\
            & (0.690, 0.856) & (0.700, 0.849) & (0.671, 0.830) & (0.668, 0.823) \\
\multirow{2}{*}{Radiologist A} &  0.733 & 0.742 & 0.678 & 0.676 \\
            & (0.648, 0.821) & (0.648, 0.806) & (0.602, 0.762) & (0.604, 0.764) \\
\multirow{2}{*}{Radiologist B} &  0.751 & 0.706 & 0.684 & 0.692 \\
            & (0.650, 0.827) & (0.639, 0.802) & (0.609, 0.770) & (0.613, 0.768) \\
\multirow{2}{*}{Radiologist A + B} &  0.733 & 0.733 & 0.694 & 0.697 \\
             & (0.651, 0.737) & (0.648, 0.728) & (0.610, 0.690) & (0.613, 0.691) \\
\midrule

 & \multicolumn{4}{c}{AUC of Intubation Prediction} \\ \cmidrule{2-5} 
 & 24 hours & 48 hours & 72 hours & 96 hours \\ \midrule
\multirow{2}{*}{DenseNet MoCo PT + Transformer} & 0.768 & 0.717 & 0.704 & 0.690 \\
            & (0.661, 0.874) & (0.620, 0.813) & (0.606, 0.802) & (0.595, 0.784) \\
\multirow{2}{*}{Radiologist A} &  0.828 & 0.763 & 0.726 & 0.704 \\
            & (0.747, 0.883) & (0.673, 0.831) & (0.648, 0.815) & (0.614, 0.776) \\
\multirow{2}{*}{Radiologist B} &  \textbf{0.844} & 0.752 & 0.735 & 0.704 \\
            & (0.755, 0.887) & (0.658, 0.824) & (0.633, 0.809) & (0.609, 0.771) \\
\multirow{2}{*}{Radiologist A + B} &  0.832 & \textbf{0.776} & \textbf{0.748} & \textbf{0.709} \\
            & (0.757, 0.822) & (0.668, 0.750) & (0.642, 0.729) & (0.612, 0.693) \\
\midrule

 & \multicolumn{4}{c}{AUC of Mortality Prediction} \\ \cmidrule{2-5} 
 & 24 hours & 48 hours & 72 hours & 96 hours \\ \midrule
\multirow{2}{*}{DenseNet MoCo PT + Transformer} & \textbf{0.785} & \textbf{0.776} & \textbf{0.817} & \textbf{0.814} \\
            & (0.683, 0.888) & (0.679, 0.872) & (0.746, 0.887) & (0.750, 0.879) \\
\multirow{2}{*}{Radiologist A} &  0.673 & 0.658 & 0.642 & 0.643 \\
            & (0.530, 0.811) & (0.549, 0.772) & (0.554, 0.750) & (0.547, 0.729) \\
\multirow{2}{*}{Radiologist B} &  0.590 & 0.576 & 0.589 & 0.631 \\
            & (0.429, 0.722) & (0.443, 0.697) & (0.488, 0.695) & (0.546, 0.728) \\
\multirow{2}{*}{Radiologist A + B} &  0.626 & 0.625 & 0.627 & 0.640 \\
            & (0.489, 0.630) & (0.507, 0.625) & (0.530, 0.630) & (0.552, 0.642) \\
\bottomrule
\end{tabular}
\label{tab:reader_study_extended}
\end{table}
The extended results for the reader study are shown in Table \ref{tab:reader_study_extended}.
In this case we bold the best-performing method when the test AUC of all other methods are outside its 95\% Bootstrap confidence interval.
The DenseNet MoCo PT + Transformer models had the highest test set AUC across all tasks except intubation and ``Any'' adverse event at 24 hours.
Generally, these scores occurred with significant overlap among the bootstrap confidence intervals.
The one setting where we observed significance was in the case of mortality prediction, where MIP model AUCs were substantially higher than that of radiologists across all time points.

\clearpage

\bibliographystyle{unsrt}
\bibliography{references}

\end{document}